\documentclass[10pt, conference]{IEEEtran}

\usepackage{amsmath}
\usepackage{graphicx}
\usepackage{float}
\usepackage{booktabs}

\begin{document}
%
\title{Efficient and Invariant Convolutional Neural Networks for Dense Prediction}



\author{\IEEEauthorblockN{Hongyang Gao}
\IEEEauthorblockA{Washington State University\\
Pullman, WA 99164\\
Email: hongyang.gao@wsu.edu}
\and
\IEEEauthorblockN{Shuiwang Ji}
\IEEEauthorblockA{Washington State University\\
Pullman, WA 99164\\
Email: sji@eecs.wsu.edu}}

%


\maketitle

\begin{abstract}
Convolutional neural networks have shown great success on feature
extraction from raw input data such as images.
Although convolutional neural networks are invariant to
translations on the inputs, they are not invariant to other
transformations, including rotation and flip.
Recent attempts have been made to incorporate more invariance in
image recognition applications, but they are not
applicable to dense prediction tasks, such as image segmentation.
In this paper, we propose
a set of methods based on kernel rotation and flip to enable
rotation and flip invariance in convolutional neural networks. The
kernel rotation can be achieved on kernels of 3 $\times$ 3, while kernel
flip can be applied on kernels of any size. By rotating in eight or
four angles, the convolutional layers could produce the
corresponding number of feature maps based on eight or four
different kernels. By using flip, the convolution layer can produce
three feature maps. By combining produced feature maps using maxout,
the resource requirement could be significantly reduced
while still retain the
invariance properties. Experimental results demonstrate that the
proposed methods can achieve various invariance at reasonable
resource requirements in terms of both memory and time.
\end{abstract}


%
\IEEEpeerreviewmaketitle

\section{Introduction}

With the development of high-performance hardware like GPU, deep
learning has shown its great success in solving challenging problem
in machine learning and artificial intelligence. Traditional machine
learning techniques are usually based on handcrafted features, which
would involve significant amount of engineering work and
domain-specific prior knowledge. Deep learning techniques can
automatically extract features from raw inputs and perform various
tasks such as classification and
segmentation~\cite{GoodfellowDLBook}.

Deep convolutional neural networks (CNNs)~\cite{LeCun:PIEEE,Alex12}
is a type of deep learning techniques that have achieved practical
success on a variety of tasks, including image
recognition~\cite{LeCun:PIEEE,Alex12}, segmentation~\cite{fcn15},
and reinforcement learning. One appealing property of CNNs is that
they are invariant to translations on the inputs, making them
particularly suitable for handling structured data like images and
audio signals. However, such models are not invariant to other
common transformations, such as rotation and flip. To cope with
these problems, we usually use data augmentation to increase the
number of training examples. Although this technique can achieve
some level of invariance, it cannot guarantee rotation and flip
invariance, since rotated copies of the same input usually lead to
different outputs during prediction. Also, data augmentation is
usually time-consuming as the training algorithm may take longer to
reach convergence.

In order to address the above challenges, a few recent studies have
attempted to construct CNNs that are invariant to generic
transformations such as rotation and flip in image recognition
applications \cite{group16}. These models achieve transformation
invariance by generating and combining transformed copies of the
same feature map. Due to their excessive resource requirement, these
approaches are not readily applicable to dense prediction tasks,
such as segmentation.

In this paper, we propose a set of techniques to construct efficient
and invariant CNNs for dense prediction tasks \cite{dense16}. The
proposed solution includes rotating or flipping the kernels used in
convolutional layers. It has been shown that rotating or flipping of
kernels will have the same effect as rotating or flipping of feature
maps \cite{group16}.
By rotating and flipping the kernels in convolutional layer, the model
will produce feature maps that can guarantee rotation and flip
invariance. A key observation is
that na\"{\i}ve implementation of many rotation and flip operations
would lead to prohibitive memory requirement on segmentation
problems. To address this challenge, we propose to use Maxout to
reduce the resource requirement while still retain the invariance
properties. Experimental
results demonstrate that the proposed methods can achieve various
invariance at reasonable resource requirements in terms of both
memory and time. Note that, although the proposed methods are mainly
described in the context of image segmentation in the following, the
proposed methods are generically applicable to dense prediction
tasks.

\section{Background and Related Work}\label{sec:RW}

Convolutional layers are not invariant to some image transformations
such as flip and rotation. For a trained CNN, it may produce a
different prediction if the input image is flipped. There are mainly
two methods to improve the invariance of CNNs; those are, dealing
with training examples, which is also called data augmentation, and
incorporating image transformation invariance within the network.

\subsection{Data Augmentation}

The key idea of data augmentation is to produce transformed training
examples and feed them to the convolutional neural networks. By
learning from these transformed images, the networks would also
learn transformed features. The only work needed for data
augmentation is to generate additional transformed training
examples. Due to its simplicity, data augmentation has been widely
used during the training process of various neural networks, such as
AlexNet \cite{Alex12} and VGG net \cite{simonyan2015very}. Simple
data augmentation operations such as flip have even been integrated
into many deep neural network tools. Although data augmentation is
effective and easy to implement, it has several drawbacks.
Specifically, data augmentation requires hand engineering on
training examples and needs more training iterations for
convergence. These additional costs reduce the efficiency of deep
models during both training and testing.

\subsection{Invariant CNNs for Image Recognition}

To overcome the limitation of data augmentation, some attempts have
been made to incorporate transformation invariance into
convolutional neural networks. Dieleman \emph{et al.}
\cite{rotation15} proposed to generate various rotated and flipped
images and feed them into the same convolutional layer. The outputs
of these layers are concatenated and fed to the following layers.
This can deal with rotation and flip to some extent, but it still
suffers from the tedious work of data augmentation. Dieleman
\emph{et al.} \cite{cyclic16} proposed a framework that rotates
feature maps within the network to achieve rotation invariance.

Cohen and Welling \cite{group16} studied rotation invariance and
equivariance of convolutional neural networks from the perspective
of group theory. They show that by operating on feature maps or
kernels, the network can achieve invariance and equivariance. Both
Dieleman \emph{et al.} \cite{cyclic16} and Cohen and Welling
\cite{group16} show that rotating feature maps and rotating kernels
within convolutional layers have the same effect.

\section{Efficient and Invariant CNNs for Dense Prediction}

\begin{figure}[t]
\centering
\includegraphics[width=0.8\columnwidth]{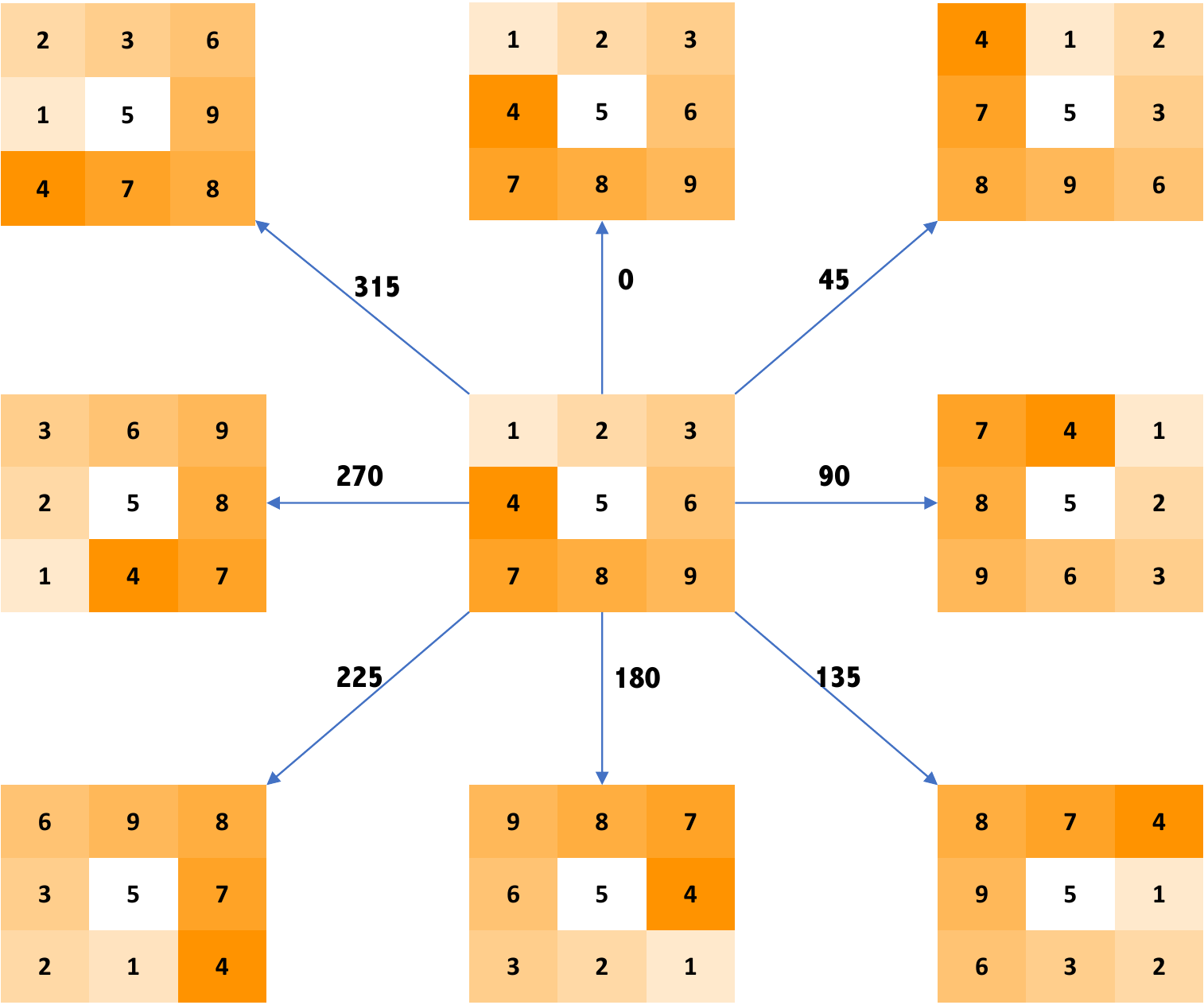}
\caption{Illustration of how a $3\times 3$ convolutional kernel can
be rotated by eight different angles to generate eight kernels.}
\label{fig:kernelrotate}
\end{figure}

\begin{figure}[t]
\center
\includegraphics[width=0.9\columnwidth]{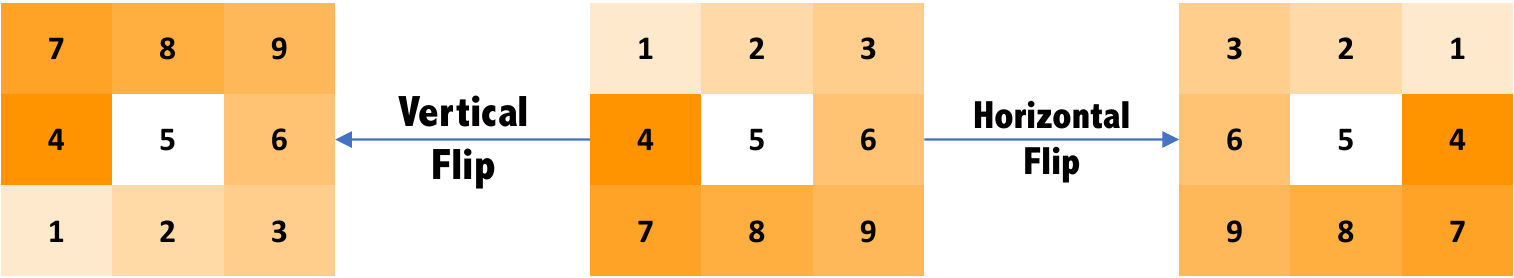}
\caption{Illustration of how a $3\times 3$ convolutional kernel can
be flipped horizontally and vertically.} \label{fig:kernelflip}
\end{figure}

\subsection{Dense Prediction}

Dense prediction problems appear frequently in image-related
applications like depth prediction~\cite{eigen2014depth}, and
image-to-image translation~\cite{laina2016deeper}. In this work, we
focus on image semantic segmentation. Long \emph{et al.}
\cite{fcn15} proposed the fully convolutional network (FCN) for
semantic segmentation. FCN employs a skip architecture to integrate
semantic information from different layers, which helps to maintain
low level features and spatial information for final dense
prediction. Based on FCN, Ronneberger \emph{et al.} \cite{unet15}
proposed the U-net, which uses symmetric encoder and decoder
architecture. The encoder part encodes the input images into
high-level feature maps by convolutional layers and max pooling
layers. The decoder part restores it to segmentation maps of the
same size as input image by up-convolution layers. In order to
maintain low-level features and spatial information, skip
connections are used in U-net. Another framework that deals with
semantic segmentation is the deconvolution network proposed by Noh
\emph{et al.} \cite{deconv15}. This framework overcomes the
limitations in FCN-based frameworks; namely fixed-size receptive
field and over-simple deconvolution procedure. For common dense
prediction tasks, data augmentation is also used to improve their
performance.

\subsection{Kernel Rotation and Flip}

In this paper, the rotation and flip invariance is achieved by
rotating and flipping kernels. Although kernel rotation achieves the
same effect as feature map rotation, it has various advantages.
First, kernel operations would not require the feature maps to be in
square shape. Second, kernel operations would be more flexible. For
example, kernels could be rotated by various angles. In our new
convolutional layers, kernels with size 3 $\times$ 3 could be
rotated by 8 different angles: 0, 45, 90, 135, 180, 225, 270, and
315 degrees. Third, the number of parameters within the model would
be reduced. For eight output feature maps in the new convolutional
layers, they are produced by eight kernels with the same weights.
The reduction of parameters could help to avoid the issue of
over-fitting, especially for small training dataset.

In the new convolutional layer, kernel rotation is performed on 3
$\times$ 3 kernels. It has been shown that 3 $\times$ 3 kernels
could achieve the effect of larger kernel sizes, such as 5 $\times$
5 and 7 $\times$ 7, with even less parameters by stacking multiple
convolutional layers. For VGG net, the kernel sizes of its
convolutional layers are mostly 3 $\times$ 3. As 3 $\times$ 3
kernels could be rotated by eight angles, the kernel rotation could
be implemented by at most 8 angles of 0, 45, 90, 135, 180, 225, 270
and 315 degrees. All eight rotated kernels will share the same set
of parameters. For the new convolutional layer, kernels could be
rotated by at most 8 different angles. The network could also choose
to rotate kernels by 4 angles, which provides more flexibility for
neural network construction. Trade-offs could be achieved between
the number of rotated kernels and the number of output feature maps.
In the experiment, the kernels in our convolutional layers are
rotated by four angles: 0, 90, 180, 270 degrees. The number of
output feature maps is doubled compared to the original
convolutional layer. This will still save half of the parameters in
the modified convolutional layers.

In the new convolutional layer, flip operations are also allowed.
Kernels could be flipped both horizontally and vertically. This will
produce three flipped versions of kernels within the new
convolutional layer. Figure \ref{fig:kernelflip} shows how flip
operations are implemented on kernels. Flipping kernel could be applied to
kernels of any size. The number of output feature maps
will only increase by three times when holding the number of
parameters in the new convolutional layer the same as previous
convolutional layer. In the experiment, the number of output feature
maps for our new flip invariance convolutional layer increases to
three times in order to maintain the same number of parameters.

\subsection{Efficient Invariance via Maxout}

\begin{figure}[t]
\includegraphics[width=\columnwidth]{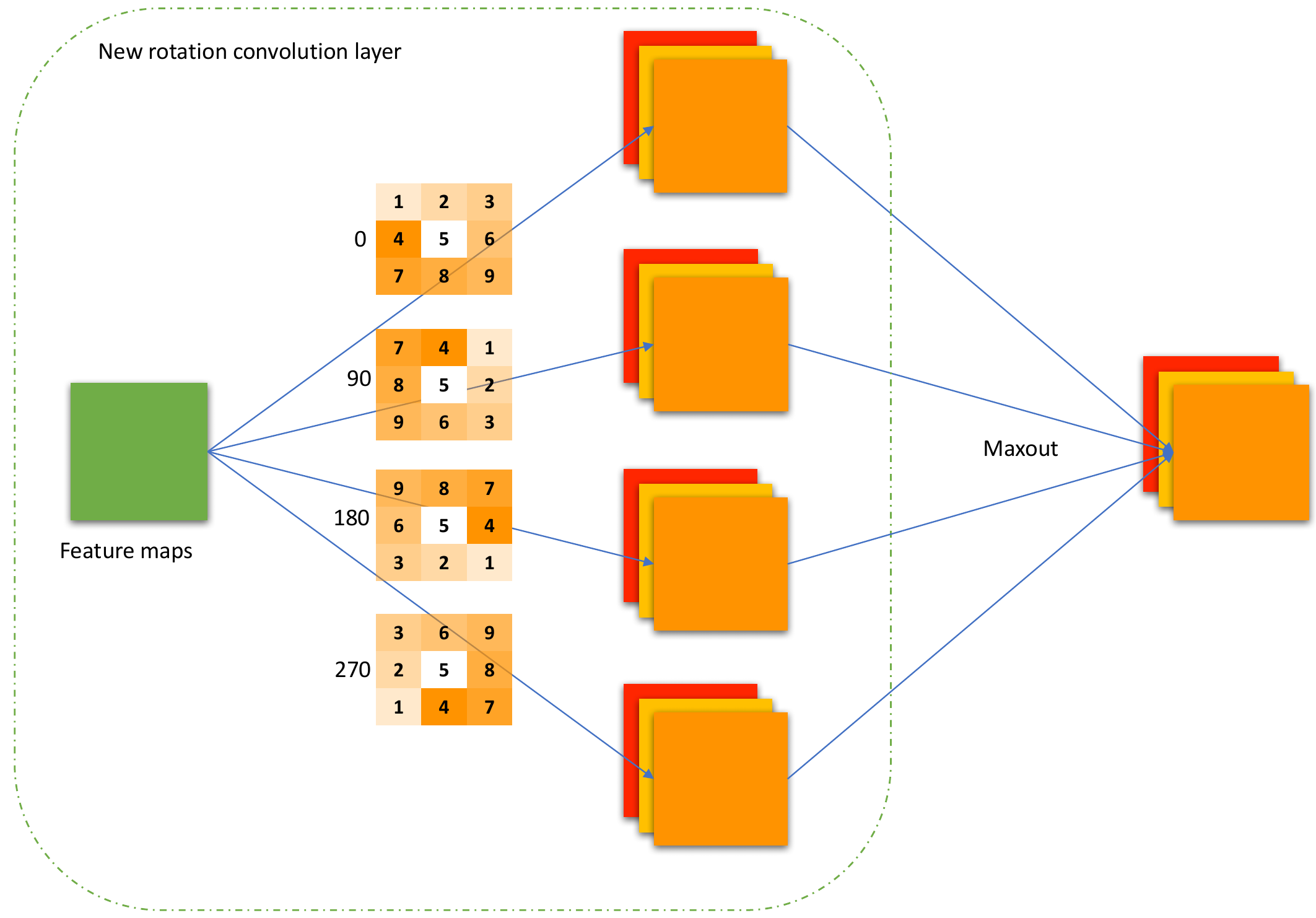}
\caption{Illustration of how to combine rotation invariance
convolutional layer and maxout to reduce the number of output
feature maps.} \label{fig:maxoutrotate}
\end{figure}

\begin{figure}[t]
\includegraphics[width=\columnwidth]{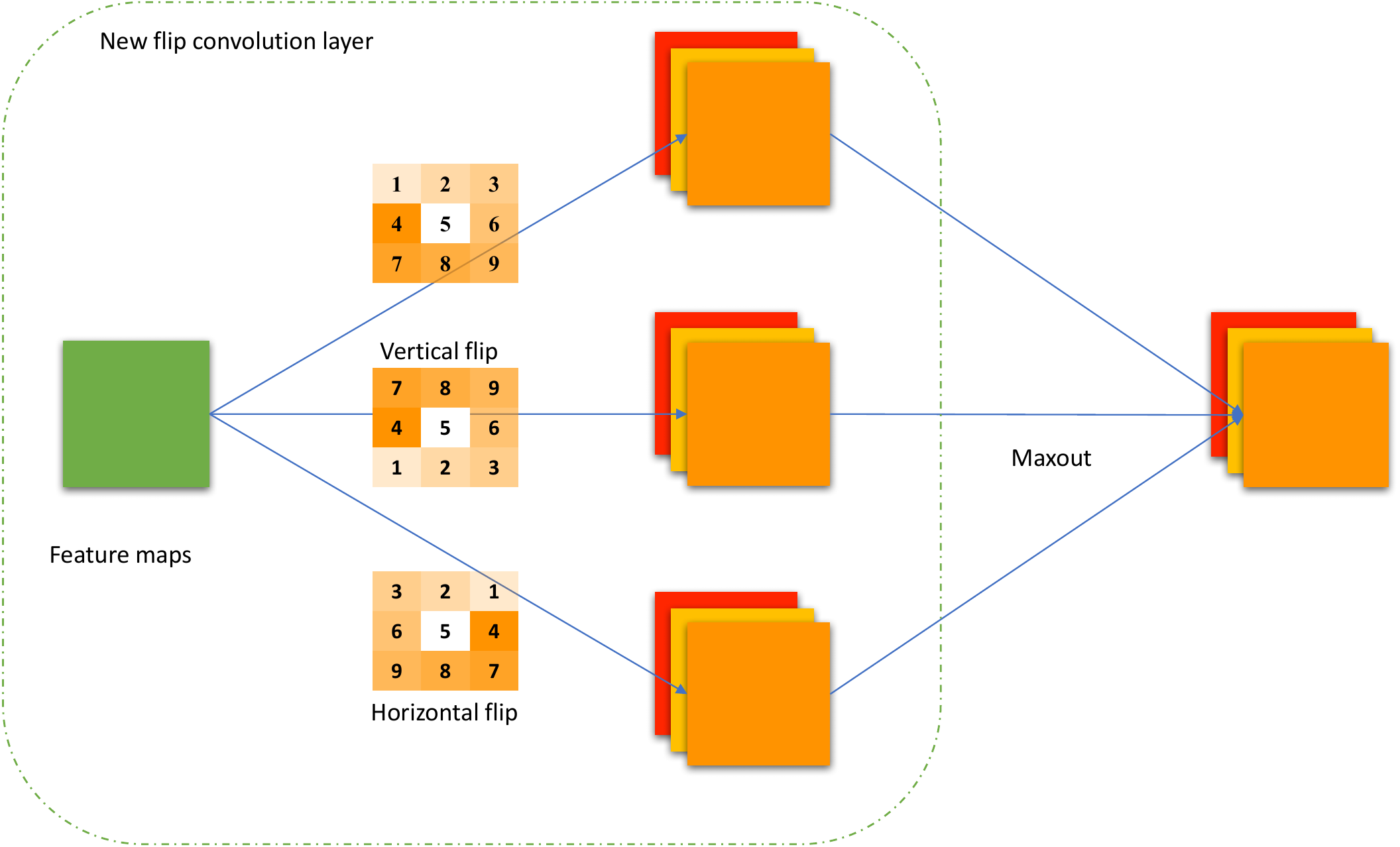}
\caption{Illustration of how to combine flip invariance
convolutional layer and maxout to reduce the number of output
feature maps.} \label{fig:maxoutflip}
\end{figure}

In practice, the number of output feature maps of the new
convolutional layer will be large in order to offset the effect of
large reduction in the number of parameters in convolutional layers
by parameter sharing. At the same time, the following convolutional
layers would receive a large number of input feature maps. For these
who want to use rotation and flip invariance convolutional layers,
they may face the problem of memory requirement in the following
convolution layers since the number of input feature maps increases
significantly. In order to cope with this problem and also enable
the flexibility of parameter numbers, we propose to use the maxout
operation \cite{maxout13} in the new convolutional layer. The maxout
networks, proposed by Goodfellow et al. \cite{maxout13}, was
designed to facilitate dropout and improve model performance. The
key idea of maxout network is to output the max of several input
feature maps. When applied to several feature maps with the same
size, maxout will produce a single feature map of the same size. In
this work, maxout is integrated with rotation and flip convolution
layers to achieve rotation and flip invariance. Figures
\ref{fig:maxoutrotate} and \ref{fig:maxoutflip} show how maxout is
integrated with rotation and flip invariant convolutional layers.

\begin{figure*}[t]
\centering
\includegraphics[width=\textwidth]{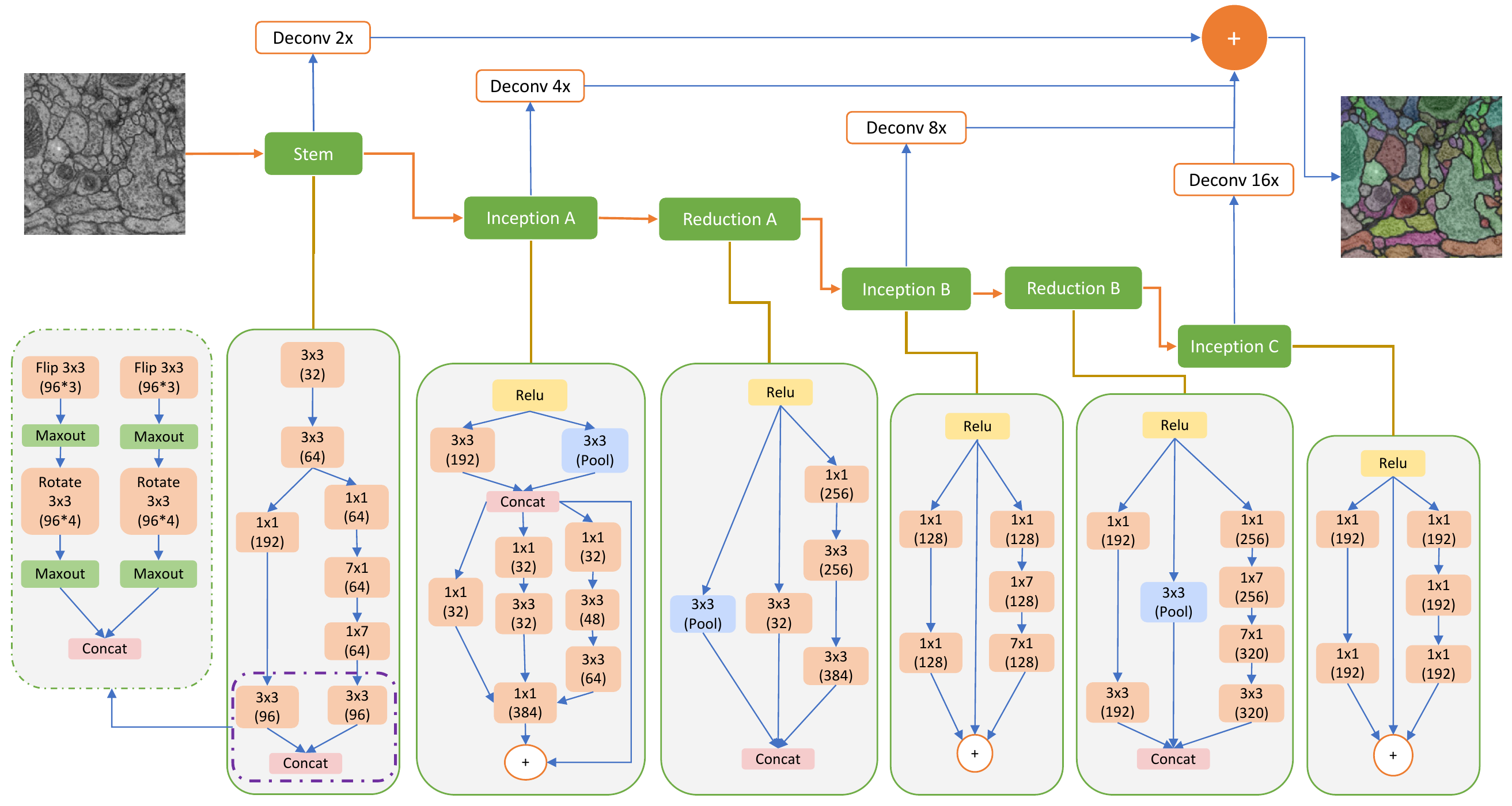}
\caption{The convolutional neural network architecture used in this
work. The two convolutional layers with 3 $\times$ 3 kernel size in
the Stem part are replaced by the new convolutional layers to
achieve rotation and flip invariance.} \label{fig:model}
\end{figure*}

When flipping kernels within the new convolutional layers, the
number of output feature maps will be tripled. The output feature
maps can be divided into three parts corresponding to three kernels.
The max operation will be performed on three feature maps. The
output of maxout operation will produce the max value of these three
set of feature maps. In this way, the number of input feature maps
for the following layers remains the same. The same operation could
be applied to rotation invariance convolutional layer. With the help
of maxout operation, the new convolutional layers become more
flexible when integrated into existing convolutional layers.

\section{Experimental Studies}\label{sec:exp}

\subsection{Dataset}

We use the dataset from the open challenge on Circuit Reconstruction
from Electron Microscopy Images (CREMI)\footnote{https://cremi.org/}
to evaluate the proposed invariant CNNs. The CREMI dataset consists
of brain electron microscopy images (EM), and the ultimate goal is
to reconstruct neurons at the micro-scale level. A critical first
step in neuron reconstruction is to segment the EM images. The main
objective of neuron segmentation is to distinguish different neuron
objects in the electron microscopy images. A common way for
segmenting EM images is to predict the neuronal boundaries in the
images~\cite{Fakhry:TMI}. For each pixel in the dense prediction
output, there will be two labels with corresponding probabilities.
Class 1 pixels correspond to boundaries in the image, and class 0
pixels correspond to all other structures. This task has an
additional problem of imbalanced samples since there are far less
boundary pixels in the image than non-boundary pixels. Thus, the
commonly-used accuracy metric may not correctly reflect the
performance of dense prediction models. In this situation, the ROC
curve is used to evaluate models involved in this work, which could
avoid the influence of imbalanced labels.

\begin{figure}[t]
\center
\includegraphics[width=0.9\columnwidth]{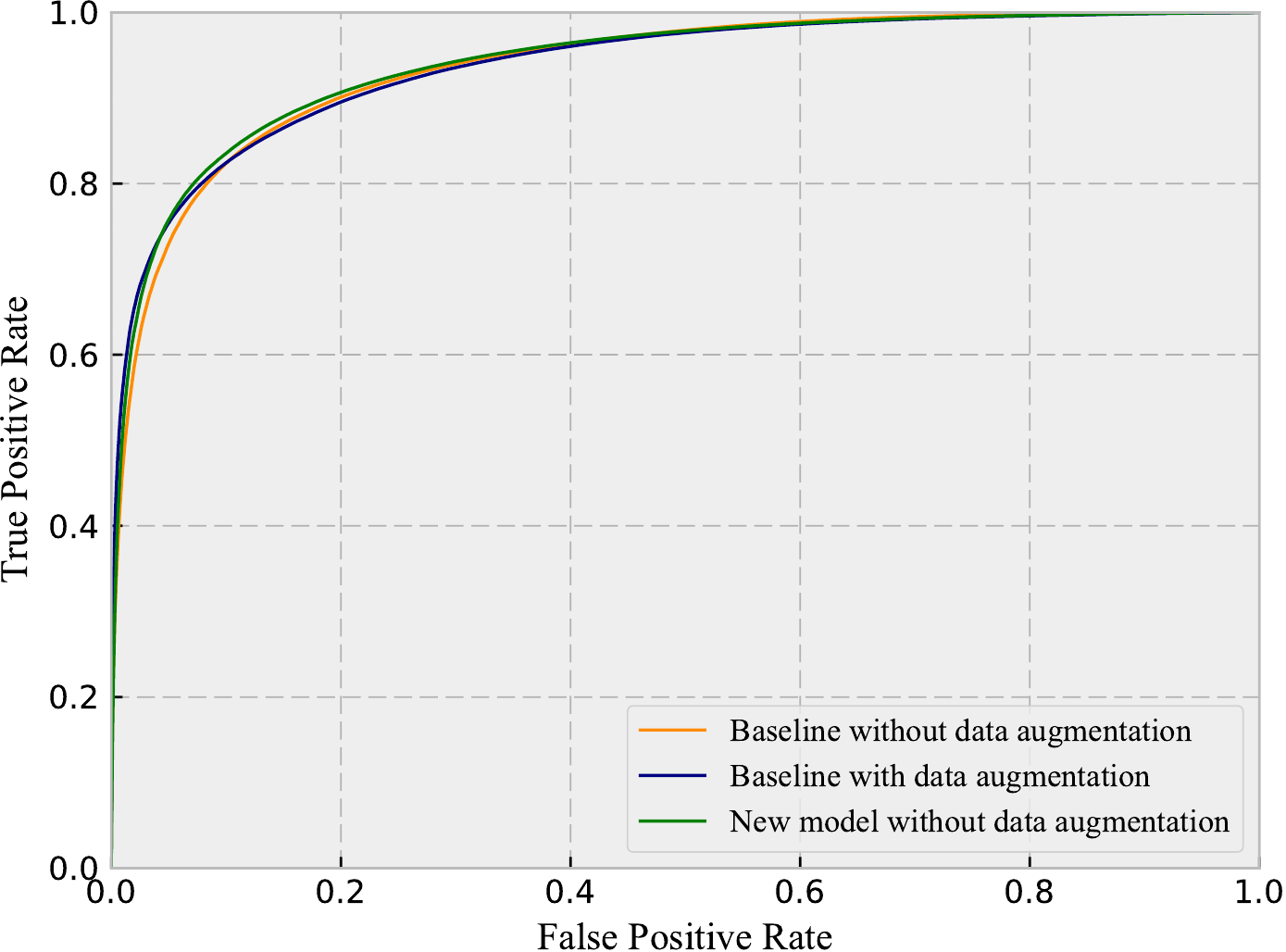}
\caption{Comparison of the ROC curves of the three models.}
\label{fig:roc}
\end{figure}

\begin{figure}[t]
\center
\includegraphics[width=0.9\columnwidth]{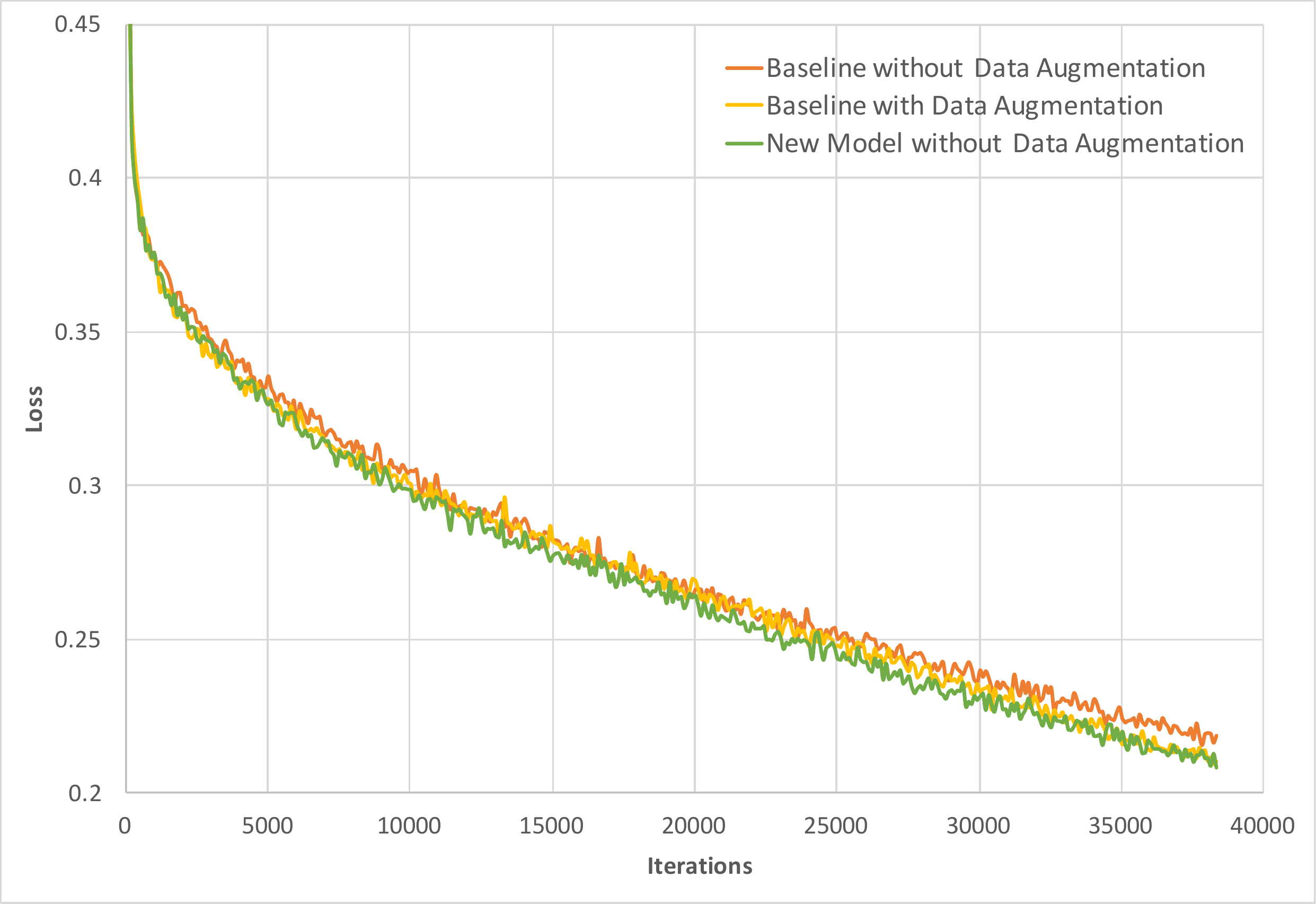}
\caption{Comparison of the training losses of the three models at
different iteration numbers.} \label{fig:loss}
\end{figure}

\subsection{Experimental Setup}

The baseline CNNs model is given in Figure \ref{fig:model}. The
baseline model consists of six modules; namely Stem, Inception A,
Reduction A, Inception B, Reduction B, and Inception C. The output
feature maps from Stem, Inception A, and Inception B are processed
by deconvolution and concatenated for final prediction. The six
modules in the model are sequential, which means the latter modules
will rely on the outputs of previous modules. As the modules
in this model are sequential, we can replace several convolutional
layers in the first module (Stem). If the rotation and flip
invariance could be achieved in this module, it can be achieved in
the following modules.

\begin{figure*}[ht]
\begin{center}{\scriptsize{
$\begin{array}{cccc}
\includegraphics[width=0.47\columnwidth]{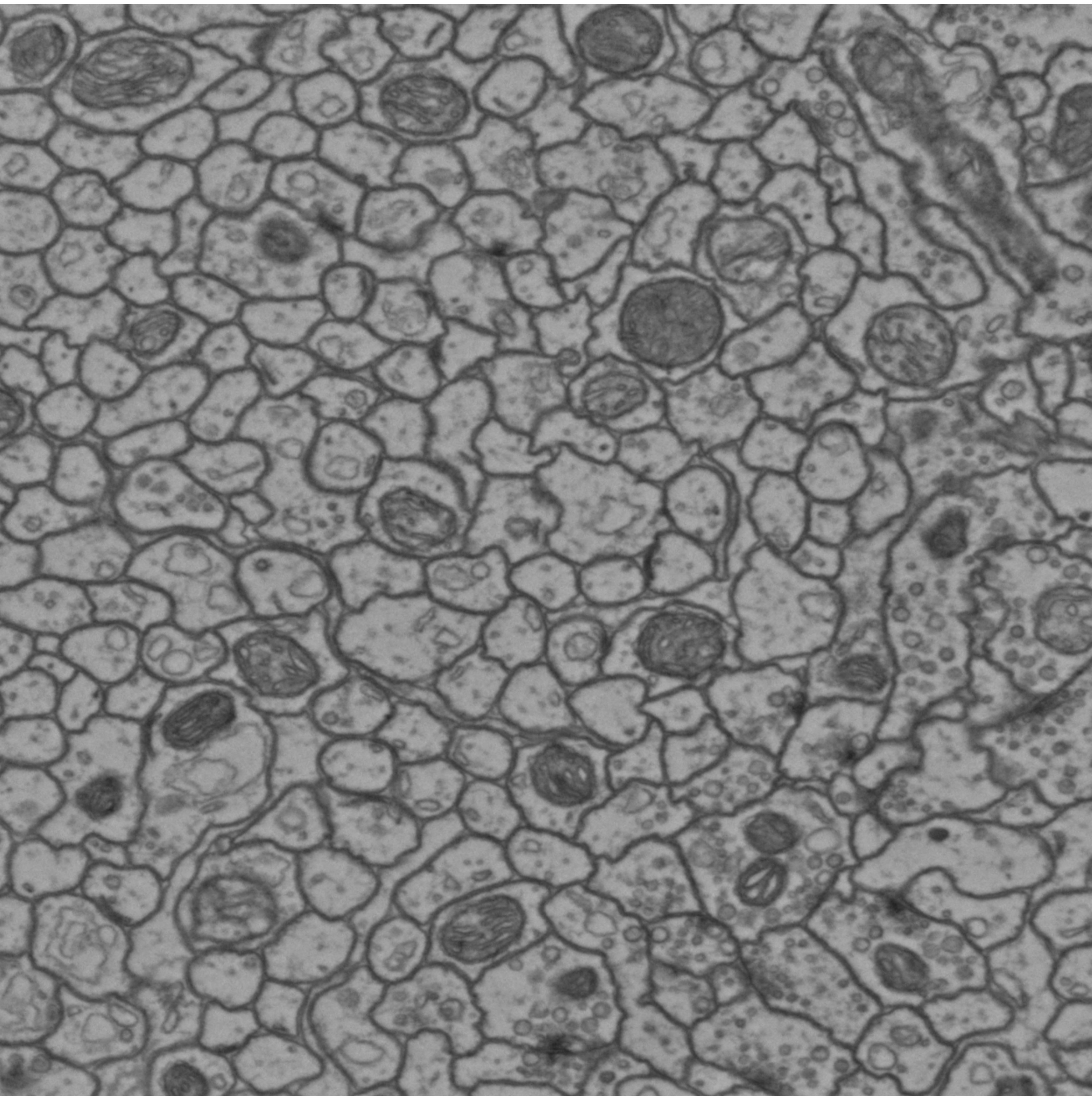}&
\includegraphics[width=0.47\columnwidth]{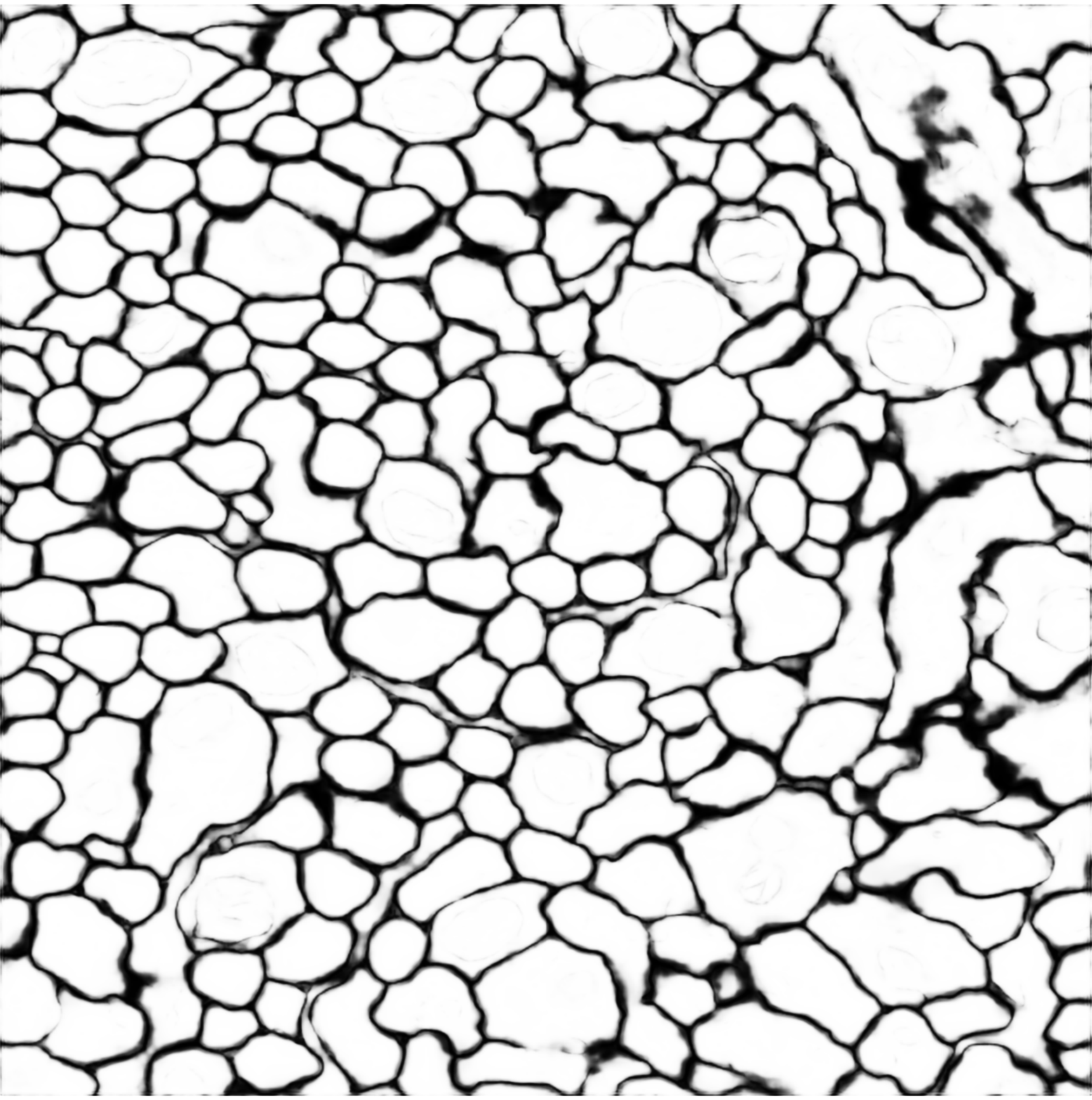}&
\includegraphics[width=0.47\columnwidth]{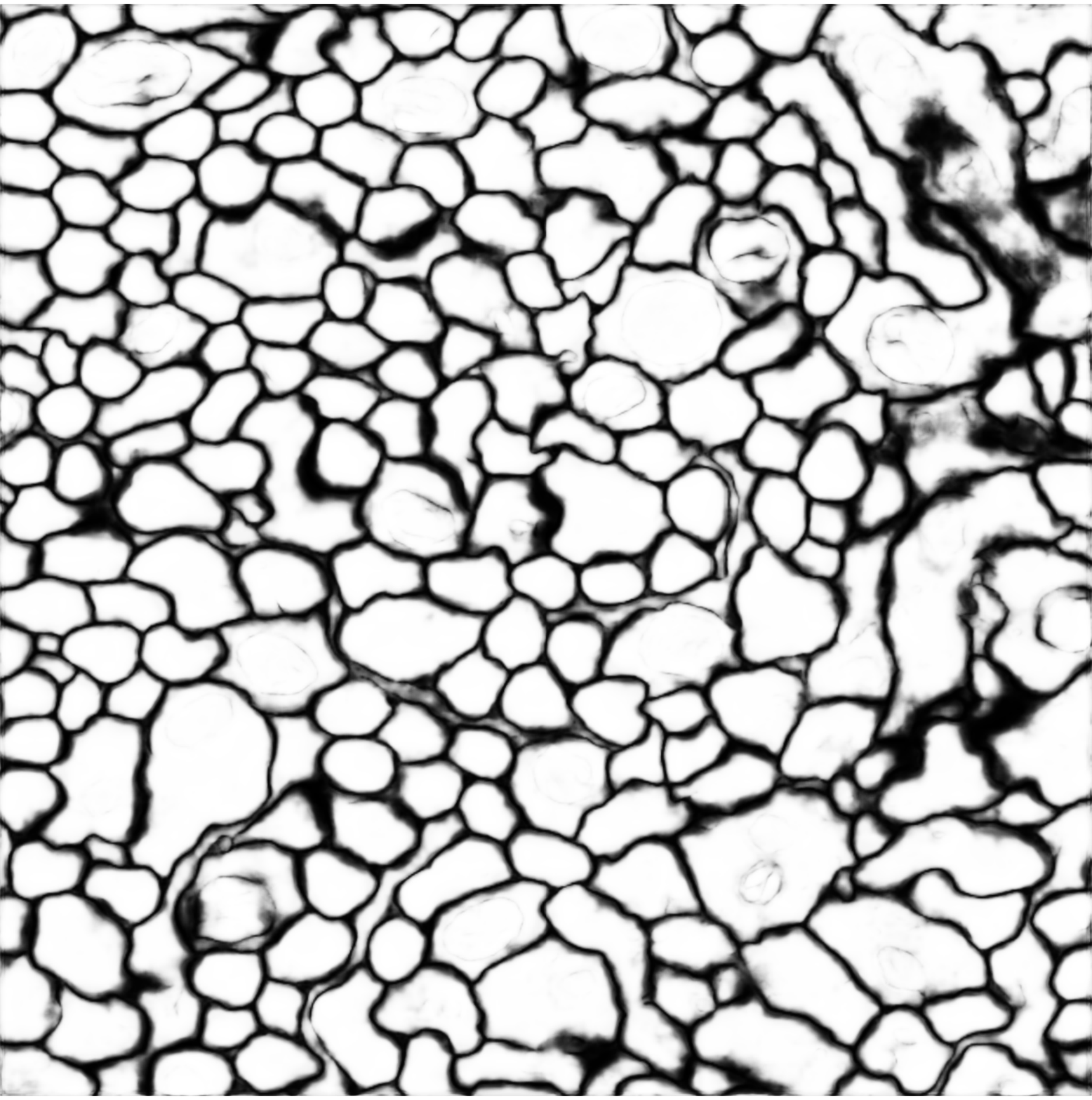}&
\includegraphics[width=0.47\columnwidth]{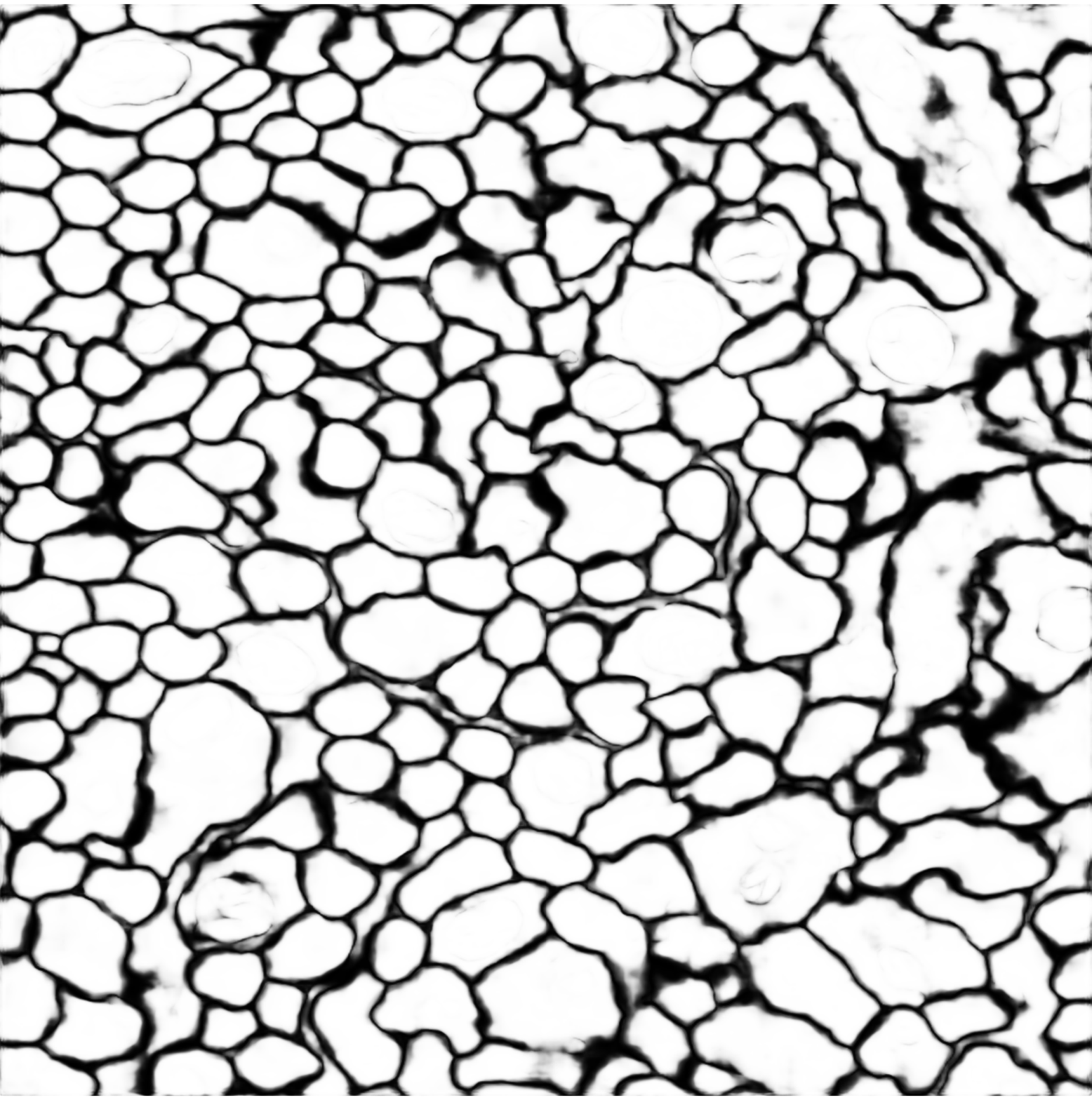} \\
\mbox{Input image}& \mbox{Baseline without data augmentation}&
\mbox{Baseline with data augmentation}& \mbox{New model without data
augmentation}
\end{array}$}}
\end{center}
\caption{Comparison of the results generated by the three models.
The inputs are brain electron microscopy images, and the outputs are
boundaries of neurons.} \label{fig:output}
\end{figure*}

\subsection{Deep Model Architectures}

As in Figure \ref{fig:model}, the two original convolutional layers
in the Stem module are replaced by the new convolutional layers. By
achieving rotation and flip invariance in these two convolutional
layers, both the outputs of the Stem module and outputs of the
following modules would be rotation and flip invariant. The kernel
sizes of these two convolutional layers are both 3 $\times$ 3. The
two original convolutional layers are replaced by two new
convolutional layers; namely rotation invariant convolutional layer
and flip invariant convolutional layer. In order to maintain the
same performance of convolutional layers, the number of parameters
in new convolutional layers are the same as before. The output
features of flip invariant convolutional layer are tripled. The
kernels in rotation invariant convolutional layer are rotated at
angles of 0, 90, 180, and 270 degrees whose output features are
quadrupled. To reduce the number of feature maps to the original
number, two maxout layers are applied to ensure the use of the new
invariant layers will have no influence on other parts of the model.
The application of the new convolutional layers in this model would
incur increased usage of memory, and the efficient invariance
through maxout makes it affordable in most situations.

In this experiment, the baseline models with and without data
augmentation are compared with the new model without data
augmentation. For baseline model with data augmentation, the
training images are augmented through a combination of rotation and
flip operations. When making predictions,
the input image will be processed with the same
augmentation operations.
The final prediction is based on majority voting
of all dense prediction outputs. The model with data
augmentation could also benefit from this ensemble strategy.
However, since the prediction of the baseline model with data
augmentation will produce more outputs, the total prediction time
would be much longer than that of the baseline model without
data augmentation. The model with the new convolutional layers does
not need any data augmentations. The prediction process will
be the same as the baseline model without data augmentation. It is
anticipated that the prediction time of the new model will be
slightly higher than that of the baseline model without data
augmentation, since there are more layers and feature map outputs in
the new convolutional layers. However, its prediction time would be
much shorter than that of baseline model with data augmentation.

\subsection{Flexibility on Rotation Angles}

In this experiment, the kernels in the new rotation invariant
convolutional layers are rotated by four angles: 0, 90, 180, and 270
degrees. In practice, the rotation angles could be more flexible.
Usually, more rotation angles means more memory requirements. If
memory resources are limited, the rotation angles could be reduced
to 0 and 180. If better rotation invariant features are needed, the
rotation angles could be increased to 0, 45, 90, 135, 180, 225, 270,
and 315 degrees. The combination of maxout with the new
convolutional layers could ensure that the number of final output
feature maps is the same as the original layers. When using the new
convolutional layers, there is great flexibility on rotation angles
based on practical needs.

\subsection{Analysis of Results}

The dense prediction outputs of three models are given in Figure
\ref{fig:output}. Figure \ref{fig:loss} shows the training loss of
the three models at different iterations. The training loss of the
baseline model without data augmentation is the highest among the
three models. The new model without data augmentation even has the
lowest training loss along the whole training iterations. This means
the property of rotation and flip invariance in the new model help
improve the training, since it could deal with features with
arbitrary rotation and flip transformations. The baseline model
without data augmentation does not have this invariance capability.
The baseline model with data augmentation may be trained to learn
this property but with much more iterations. From these results, it
is clear that the baseline model with data augmentation has larger
loss than the new model until after 35,000 iterations. But the loss
of the baseline model without data augmentation is always higher
than that of the new model. The gap between the loss curves of the
baseline model without data augmentation and the new model even
increases as iteration number increases.

Table \ref{table:auc} shows the model evaluation results using AUC
values. Since the prediction output labels are imbalance, AUC values
would be more meaningful in the evaluation. The AUC values of the
new model and baseline model with data augmentation are similar and
higher than that of the baseline model without data augmentation.
This shows that the new model has very similar performance compared
to the baseline model with data augmentation. This also demonstrates
that data augmentation usually leads to improved performance, a
results that is consistent with prior
observations~\cite{Fakhry:bioinfo16}. Figure \ref{fig:roc} gives the
ROC curves of three models.

Table \ref{table:time} provides the prediction time for the three
models. The baseline model without data augmentation has the highest
prediction speed. The prediction time for the new model without data
augmentation is slower than that of the baseline model without data
augmentation. This is because there are more convolutional layers
and more output feature maps in the new convolutional layers. As the
replaced part is at the early stage of the model, the size of
feature maps is relatively large, which could increase the
computational time. The baseline model with data augmentation has
the highest prediction time. It is far longer than that of the new
model. The prediction process of the baseline model with data
augmentation will produce eight dense prediction outputs
corresponding to eight data augmentation operations in the training
process. Thus, the total prediction time of the baseline model with
data augmentation is eight times of that of the baseline model
without data augmentation. Under the same prediction performance,
less prediction time should be appreciated, which makes it more
applicable in practice. Overall, the experimental results show that
the proposed invariant methods achieve higher performance as
compared to the original method without data augmentation. In
addition, the proposed methods are very efficient when making
predictions, as compared to the original methods with data
augmentation.

\begin{table}[t]
\centering \caption{Comparison of AUC values achieved by the three
models.} \label{table:auc}
\begin{tabular}{  l  | c  }
    \hline
    \textbf{Model} & \textbf{AUC values}      \\ \hline\hline
    Baseline without Data Augmentation      & 0.931  \\ \hline
    Baseline with Data Augmentation        & 0.939  \\ \hline
    New Model without Data Augmentation     & 0.941  \\
    \hline
\end{tabular}
\end{table}

\section{Conclusion and Discussion}\label{sec:conclusion}

In this work, we proposed a new convolutional layer that could
achieve rotation and flip invariance. The new convolutional layer is
implemented through kernel operations instead of feature map
operations to make it more flexible and efficient. For rotation
invariant convolutional layer, the kernels in the layer could be
rotated at 8 angles: 0, 45, 90, 135, 180, 225, 270, and 315 degrees.
The rotated angles could be reduced to 2 or 4 angles to meet memory
constraints. For flip invariant
convolutional layer, the original kernels are flipped both
horizontally and vertically. In order to reduce the
influence of increased number of feature maps from the new
convolutional layers, we propose to use
maxout to compress the feature maps.
The combination of the new convolutional layers and maxout layer
makes it even more flexible when applied to other models. In experiment,
we compared the proposed
new model with baseline models with and without data augmentation.
The experimental results show that the proposed invariant methods
achieve higher performance compared to the original method
without data augmentation.

\begin{table}[t]
\centering \caption{Comparison of the prediction time of the three
models.} \label{table:time}
\begin{tabular}{  l | c  }
    \hline
    \textbf{Model} & \textbf{Prediction Time} \\ \hline\hline
    Baseline without Data Augmentation  & 28m \\ \hline
    Baseline with Data Augmentation     & 226m \\ \hline
    New Model without Data Augmentation & 45m \\
    \hline
\end{tabular}
\end{table}





\section*{Acknowledgment}
This work was supported in part by National Science Foundation grant
DBI-1641223.




%

\bibliographystyle{./IEEEtran}
\bibliography{./deep,./sigproc}


\end{document}